\documentclass[9pt]{article}
\usepackage{spconf,amsmath,graphicx}
\usepackage{xspace}
\usepackage{enumitem}

\usepackage{url}

\newcommand{\rnnlms}{\textsc{rnnlm}s\xspace}
\newcommand{\rnnlm}{\textsc{rnnlm}\xspace}
\newcommand{\lstm}{\textsc{lstm}\xspace}
\newcommand{\ngram}{\textsc{n}-gram\xspace}
\newcommand{\rnns}{\textsc{rnn}s\xspace}
\newcommand{\tr}{\textsc{tr}\xspace}
\newcommand{\ptb}{\textsc{ptb}\xspace}
\newcommand{\wsj}{\textsc{wsj}\xspace}

\title{Knowledge Distillation for Recurrent Neural Network Language Modeling
With Trust Regularization}
\name{Yangyang Shi$^{\star }$ \qquad Mei-Yuh Hwang$^{\star }$ \qquad  Xin Lei$^{\star }$ \qquad Haoyu Sheng$^{\star \dagger}$}
\address{$^{\star}$ Mobvoi AI Lab \\
         $^{\dagger}$ Williams College}
\begin{document}
%
\maketitle
\begin{abstract}
Recurrent Neural Networks (\rnns) have dominated language modeling because of their 
  superior performance over traditional \ngram based models. In
  many applications, a large Recurrent Neural Network language model (\rnnlm)
  or an ensemble of several \rnnlms is used. These models have large memory
  footprints and require heavy computation. In this paper, we examine the effect
  of applying knowledge distillation in reducing the model size for \rnnlms. In addition,
  we propose a trust regularization method to improve the knowledge
  distillation training for \rnnlms. Using knowledge distillation with trust
  regularization, we reduce the parameter size to a third of that of the previously published best model while maintaining the state-of-the-art perplexity result on Penn Treebank data. In a speech recognition N-best rescoring task, we reduce the \rnnlm model size to $18.5\%$ of 
  the baseline system, with no degradation in word error rate (WER) performance on Wall 
  Street Journal data set.
\end{abstract}

\section{Introduction}
Recurrent Neural Networks are currently popular choices for language modeling.
Recurrent Neural Network based language models (\rnnlms) have outperformed
traditional N-gram based language models in many machine learning
tasks, such as automatic speech recognition~\cite{mikolov2010},
machine translation~\cite{Schwenk2012} or text summarization~\cite{Rush2015}.

To address the data sparsity issue in \ngram language models,
\rnnlms represent each word in a continuous and lower dimensional space.
In contrast to \ngram language models which only capture local short distance
dependencies, \rnnlms are capable of representing full histories of variable lengths
with recurrent vector representations.

However, in many natural language processing tasks, large models are usually required to achieve state-of-the-art performance. In deep model training, significant redundant representations are learned~\cite{Denil2013}. Due to their large memory footprint and daunting computational cost, the applications of large networks in practical scenarios are greatly hindered.
To reduce the deep neural network model size, model quantization~\cite{Lin2016},
model weight pruning~\cite{See2016}, low rank matrix factorization~\cite{Xue2013}
and knowledge distillation~\cite{Hinton2015}
are usually applied. In this paper, we consider the knowledge distillation method
in the context of \rnnlms and their application in speech recognition N-best rescoring.

Knowledge distillation is also referred to as teacher student training, where a
small model (student) is trained to match the output of a large
model (teacher). In addition to model compression, knowledge
distillation has been used in transfer learning and domain adaptation tasks.
Depending on the specific scenario, the teacher and student models can be trained
on the same or different data~\cite{Hinton2015}. In this work, both of the teacher and student models are trained on the same data but with different model sizes. 

In knowledge distillation, the student model is learned to minimize the combination of cross-entropy
loss based on training data labels and Kullback-Leibler (KL) divergence to teacher model distribution. In this paper, our experiments in the context of language modeling
show that the knowledge distillation methods~\cite{Hinton2015} that use
cross-entropy loss and KL divergence with fixed interpolation weights get worse results than using KL divergence alone. Hence
 we propose a trust regularization (\tr)  method to dynamically adjust
the combination weights for these two types of losses. Two experiments are performed to
verify the effectiveness of the proposed method. In the first experiment, the 
student model achieves state-of-the-art perplexity results on the Penn Treebank
dataset~\cite{mikolov2010} with a model size one third of that of the
previously published 
best model. The second experiment is speech recognition N-best rescoring on Wall
Street Journal dataset~\cite{Paul1992}, where the student model size is only $18.5\%$ of that from its teacher model and yet achieves similar word error rates.

\section{Knowledge Distillation for \rnnlms}
\subsection{Teacher Model: A high-rank \rnnlm with regularizations}
In this paper, knowledge distillation is built on top of a high-rank \rnnlm~\cite{Yang2017}
with several \rnnlm specific regularizations and optimization methods~\cite{Merity2017}.

The high-rank \rnnlm uses a mixture of softmaxes (MoS) to make the softmax layer
in \rnnlm more expressive. Similar to conventional \rnnlms~\cite{Zaremba2014,Merity2017},
a sequence of hidden states is obtained after processing the input sequence $X$
over a stack of recurrent layers. On top of the hidden states, the MoS represents
the conditional distribution of current word $x_t$ 
as weighted sum of different softmax layers. 

To achieve optimal performance for \rnnlms, the following regularization
techniques~\cite{Merity2017} are applied.
\begin{itemize}[itemsep=0\baselineskip]
\item Three different dropouts: DropConnect~\cite{Wan2013} is the dropout to the weight matrices within \lstm cells for hidden to hidden transitions.
Variational dropout is applied to all inputs and outputs for \lstm cells.
Embedding dropout~\cite{Gal2015} is equivalent to using variational dropout on embedding layers.
\item Activation Regularization is applied to penalize large hidden layer activations and to penalize dramatic difference in activations across neighboring frames.
\end{itemize}
However, in contrast to cross-entropy training, to achieve the best performance for knowledge distillation, our experiments reveal that  all these regularization methods need to be turned off for student model training.

\subsection{Knowledge Distillation}
The basic idea of knowledge distillation~\cite{Hinton2015} is to train a smaller student model by
providing additional information in the form of outputs from a larger teacher model.
Usually the student model, denoted by $\theta$, is trained to minimize the
interpolation of a cross-entropy loss according to training data labels (hard labels)
and KL divergence between the student model outputs and the outputs from a teacher model (soft labels):
\begin{equation}\label{eq:loss}
  L(\theta) = \alpha L_{CE}(\theta) + ( 1-\alpha ) L_{KL}(\theta)
\end{equation}
where $L_{CE}(\theta)$ is the cross-entropy loss. In the context of language modeling, the cross-entropy loss can be
represented as
\begin{equation}\label{eq:celoss}
  L_{CE}(\theta) = -\sum_{x}1(y=x)\log P(x|c,\theta)
\end{equation}
where $y$ is the hard label and $1(y=x)$ is the indicator function.
The KL divergence of the student output distribution
$P(x|c, \theta)$ to the teacher output distribution $Q(x|c, \theta_{te})$ can be formulated as
\begin{equation}\label{eq:kl}
  L_{KL}(\theta) = -\sum_{x}Q(x|c, \theta_{te}) \log \frac{P(x|c, \theta)}{Q(x|c,\theta_{te})}.
\end{equation}
where $Q(x|c,\theta_{te})$ is the teacher model output distribution for each word $x$. $Q(x|c,\theta_{te})\log Q(x|c,\theta_{te})$  is constant for each $x$ since the teacher model $\theta_{te}$ is fixed. Therefore the minimization
of the above equation is equivalent to minimization of the following loss
\begin{equation}\label{eq:kl2}
  L_{KL}(\theta) = -\sum_{x}Q(x|c,\theta_{te})\log P(x|c,\theta).
\end{equation}
In~\cite{Hinton2015}, the output distribution $Q(x|c,\theta_{te})$ is
represented as a softmax probability with temperature $\tau \geq 1$. We have tried different temperatures $\tau\in[1,2,5,8,10]$
in our experiments. We find that $\tau=1$ gives the best performance and is used in all experiments reported here.

\subsection{Trust Regularization}
In language modeling, each training data label is represented as a degenerated data
distribution which gives all probability mass to one class. So the degenerated data distribution is localized and can be over-confident,
comparing with the teacher's probability distribution learned over the whole training data. Different from previous
observations using knowledge distillation for
acoustic modeling~\cite{Hinton2015} and image classification~\cite{Romero2014}, our
experiment results show that the student model learned by minimizing the interpolation
of cross-entropy loss and KL divergence performs worse than the student model learned by
minimizing the KL divergence alone.

Thus, we propose the following trust regularization (\tr) method to dynamically
adjust the weight (trust) for cross-entropy loss.

\begin{equation}\label{eq:trust}
  L(\theta) = R(y)L_{CE}(\theta) +  L_{KL}(\theta)
\end{equation}
where $R(y)$ is the trust regularizer that is formulated as follows:
\begin{equation}\label{eq:r}
  R(y) = -\alpha\sum_{x}{\textsc{1}}(y=x)\log(1-Q(x|c,\theta_{te}))
\end{equation}
where $\alpha>0$ is a scalar value. The more the teacher model $Q(x|c,\theta_{te})$ agrees with the hard label
$y$, the more we trust CE and therefore the more weight the CE loss will receive.

\begin{table*}[t!]
\begin{center}
\small
\begin{tabular}{l|c|c|c}
  \hline model & \#Param &Valid & Test \\ \hline
  RNN-LDA+KN-5+cache~\cite{mikolov2012context}      & 9M   & -     & 92.0 \\
  LSTM~\cite{Zaremba2014}                     & 20M  & 86.2  & 82.7 \\
  Variational LSTM medium MC~\cite{Gal2015} & 20M  &-      & 78.6 \\
  Variational LSTM large MC~\cite{Gal2015} & 66M  &-      & 73.4 \\
  Char-CNN-LSTM~\cite{kim_char_cnn_2015}            & 19M  & -     & 78.9 \\
  Pointer-Sentinel~\cite{merity_pointer_2016}       & 21M  & 72.4  & 70.9 \\
  LSTM + continuous cache pointer~\cite{grave_continuous_cache_2016}\textsuperscript{\dag}& -    & -     & 72.1  \\
  Tied variational LSTM+augmented loss~\cite{Inan2016} &24M & 75.7  &73.2 \\
Tied variational LSTM+augmented loss~\cite{Inan2016} &51M & 71.1  &68.5 \\
  Variational RHN~\cite{zilly_rhn_2016}             & 23M  &67.9   &65.4 \\
  NAS Cell~\cite{zoph_neural_arch_search_2016}      & 25M  & -     &64.0 \\
  NAS Cell~\cite{zoph_neural_arch_search_2016}      & 54M  & -     &62.4 \\
  4-layer skip connection LSTM~\cite{melis_nlm_2017}& 24M  &60.9   &58.3 \\
  AWD-LSTM finetune~\cite{merity_regularing_lstm_2017}&24M &60.0   &57.3 \\\hline
  AWD-LSTM-Mos w/o finetune~\cite{Yang2017}      & 22M  &58.1   &56.0 \\
  Ours w/o finetune                         & 7M   &57.8 &55.6 \\\hline
  AWD-LSTM-Mos finetune~\cite{Yang2017}        & 22M  &56.5   &54.4 \\
  Our teacher model (AWD-LSTM-Mos finetune $\times$ 5)           & 22M$\times$ 5&52.7 &51.4     \\
  Ours finetune                             & 7M   &{\bf{55.9}}   &{\bf{54.0}}  \\\hline
  AWD-LSTM-Mos finetune dynamic eval~\cite{Yang2017}\textsuperscript{\dag}&22M &48.33 &47.69 \\
  Ours finetune dynamic eval\textsuperscript{\dag}               & 7M   &{\bf{48.17}}  &{\bf{47.60}} \\
\hline
\end{tabular}
\end{center}
  \caption{Single model perplexity results on validation and test sets on \ptb.
  \textsuperscript{\dag} indicates using dynamic evaluation. }
\label{tab:1}
\end{table*}
\vspace{-0.6 pt}

\section{Experiments}
To evaluate the proposed \tr in knowledge distillation for
\rnnlms, we conduct experiments on two tasks. The first task is to measure the perplexity of language models.
The widely used benchmark dataset Penn Treebank (\ptb)~\cite{mikolov2010} is used. \ptb data has a
predefined $10K$ close vocabulary set. 

The second task
is to measure the WER of \rnnlms in speech recognition N-best rescoring
on  Wall Street Journal~\cite{Paul1992} (\wsj).
For acoustic training on the SI-284 training set, and language modeling training data preprocessing,
we follow the recipe in the Kaldi speech recognition toolkit~\cite{Povey2011}.
There are $37M$ words in the \wsj LM trainining data, with $200K$ of which are separated as the validation data.
The vocabulary of the \wsj language model 
is capped at $40K$ by the above Kaldi recipe. Those training words not in the $40K$ are mapped to token {\textsc{UNK}}.
Low-frequency words in the $40K$ vocabulary are mapped to a special token as {\textsc{RNN\_UNK}} (essentially a "rare-word" class) for \rnnlm training. 

\subsection{Experiments on \ptb}
In this task, the teacher model is an ensemble of five models that are initialized based on different random seeds ($[31,37,61,71,83]$).
To train each component LM in the teacher model, we closely follow the regularization, optimization and hyper-parameter tuning techniques
that are applied by ~\cite{merity_regularing_lstm_2017,Yang2017}. Each component model has 3 layers of \lstm with 960 neurons. The embedding
size is set to 280. Before the softmax layer, the \lstm output is projected down to a bottleneck layer with size 620. In the mixture of softmax
layer, 15 experts are used. Dropout rates of 0.4, 0.29 and 0.225 are used for the \lstm input, the \lstm output,
and hidden-to-hidden transition in \lstm, respectively. For the other layers, the dropout rate is set to 0.4. The number of parameters for each
component model is $22M$.

The student model is trained to minimize the combined loss with trust regularization (Eq.~\ref{eq:r}) with $\alpha=0.1$. It
has 3 layers of \lstm with hidden size 480. The dimensions of the embedding layer and the bottleneck layer are 200 and 300, respectively. The student model
uses the same number of experts in the softmax layer as the teacher model. We do not apply dropout or activation regularization in training the student model.

Table~\ref{tab:1} gives the language model perplexity  results on \ptb data. The single student model trained via the trust regularized knowledge
distillation method outperformed all the baselines with or without fine-tuning and dynamic evaluation. Using only one third (7/22) of parameters
compared with the previously published best model~\cite{Yang2017}, our student model  achieved better perplexity.

\subsubsection{Ablation Analysis}
To further verify the contribution of the trust regularized knowledge distillation method,
we conduct ablation experiments on \ptb dataset. All the ablation experiments exclude
the usage of fine-tuning and dynamic evaluation to avoid distractive factors. Hence the first row
of Table~\ref{tab:2} is copied from the first "7M" model in Table~\ref{tab:1}. Each of the following experiments in the table only changes one factor, compared to the first row.

\begin{table}[h]
\begin{center}
\small
\begin{tabular}{l|c|r}
  \hline model &Valid & Test \\ \hline
  student model  ($\tau=1.1$)       &57.8  &55.6      \\ \hline
  -cross-entropy loss ($\alpha=0$) &58.5  &56.8      \\
  -trust regularization &59.4 & 57.4    \\
  +dropout              &65.4  &62.9      \\
  +activation regularization &58.0  &55.9  \\
  -knowledge distillation& 67.8 &65.7 \\
\hline
\end{tabular}
\end{center}
  \caption{Ablation analysis on \ptb data.}
  \label{tab:2}
\end{table}

The second row turns off cross entropy loss completely. In the third row, a constant weight of 0.1 is used for CE loss 
instead of a dynamic weight assigned by \tr. The weight for KL loss remains to be 1 as shown by Eq.~\ref{eq:trust}. Comparing these two, we find that KL loss alone is better than a constant CE weight.
This demonstrates the effectiveness of the dynamic weight from \tr.

Furthermore, as indicated by the next two rows in Table~\ref{tab:2}, we find that, different from language model learning using hard labels, knowledge distillation method renders better models without dropout or activation regularization .

Finally, without using knowledge distillation (i.e. no teacher), the small model is trained with only cross-entropy loss. In such situation, the conventional regularization and optimization techniques such as dropout are applied to get the best performance 
for cross entropy training.
It shows that without knowledge distillation, the perplexity increases by $17.3\%$ and $18.2\%$ on validation and test data.

\subsubsection{Trust Regularization vs. Fixed Weight Interpolation}
Table~\ref{tab:3} gives the comparison between \tr method and fixed weight interpolation
 in knowledge distillation. Again, to avoid distraction, neither fine tuning nor dynamic evaluation is used here to acquire the results.
 The results show that in language modeling, the
fixed weight interpolation of cross-entropy loss and KL divergence renders worse models
than using KL divergence loss alone or \tr method.
\begin{table}[h]
\begin{center}
\small
\begin{tabular}{l|c|r}
  \hline model &Valid & Test \\ \hline
  student model&57.8  &55.6      \\
  0.0CE+1.0KL           &58.5  &56.8  \\
  0.1CE+0.9KL &59.5  &57.6      \\
  0.2CE+0.8KL &59.7 & 57.6   \\
  0.5CE+0.5KL &63.5  &58.2      \\
\hline
\end{tabular}
\end{center}
  \caption{Comparison of constant interpolation vs. trust regularized interpolation for combining CE loss and KL loss.}
  \label{tab:3}
\end{table}

\vspace{-20 pt}
\subsection{Experiments on \wsj}
In this experiment, the teacher is an ensemble of two models (Comp 1 and Comp 2 in Table ~\ref{tab:4}) that are initialized with random seed 17 and 31.
Each component model has one 900 dimensional embedding layer, 3 layers of \lstm that each has 1150 hidden neurons
and one bottleneck layer with 650 neurons before softmax layer. In the mixture of softmaxes, 7 experts are used.
A dropout rate of 0.4 is used for variational dropout. ConnectDrop isn't applied here.
The embedding layer dropout rate is 0.1.
In the student model, there are also one layer of embedding, 3 layers of \lstm and one bottleneck layer.
Each layer of the student model has 250 neurons. There are also 7 experts in the mixture of softmaxes.
For student model training, no dropout is applied. The trust regularization weight is set to $0.01$.

The experimental results on \wsj are listed in Table ~\ref{tab:4}. A first-pass decoding is run by using the ngram LM of $40K$ vocabulary
generated by Kaldi scripts, to generate 1000 best hypotheses
per utterance for rescoring. While compiling the decoding graph, any word in the $40K$ vocabulary will be removed if its phonetic pronunciation cannot be found or derived. We generate two sets of 1000-best hypotheses, each with a different pronunciation dictionary. BD in Table ~\ref{tab:4} means the nbest is generated from a bigger dictionary.

The student learned from \tr knowledge distillation
method has marginally better performance than the model trained based on fixed interpolation weights (CE weight 0.01). Compared to each of the mixture component models for the teacher, 
the student model could get almost identical error rates, but with only
 $12/65\approx18.5\%$ of parameters. The table also includes the performance of a pure CE training on the small model, without teachers. The 72.5 perplexity shows the importance of having a teacher.

\begin{table}[t!]
\begin{center}
\small
\begin{tabular}{l|c|c|c|r}
  \hline Model & \#Param &Valid ppl & Dev93/BD & Eval92/BD \\ \hline
  1st-pass & - & 118 & 9.02/6.30 & 6.31/3.90 \\ \hline
  RNNLM& 2012M  & 81.2 &7.44/4.82   & 4.98/2.66 \\
 Teacher & 65$\times2$M     &{\bf 45.4} & {\bf 6.18/3.80} & {\bf 4.36/2.37} \\
  Comp 1        & 65 M  &53.2 &6.30/4.10 &4.34/2.52     \\
  Comp 2        & 65 M  &53.8 &6.31/4.12 &4.51/2.60 \\
  CE only       & 12 M  &72.5 &6.53/4.41 &4.54/2.76      \\
  fixed interp. & 12 M  &55.3 &6.34/4.13 &4.43/2.61  \\
  \tr interp.   & 12 M  &54.1 &6.30/4.13 &4.36/2.57  \\
\hline
\end{tabular}
\end{center}
  \caption{Perplexity (ppl) and word error rates on \wsj data.
  "1st-pass" is the top1 output from ngram first-pass decoding.
  'BD' means the first pass decoding uses a {\bf b}ig {\bf d}ictionary for pronunciation lookup. }
\label{tab:4}
\end{table}

\section{Conclusions}
In this paper, we apply knowledge distillation for \rnnlms. The experiments on language modeling
reveal that the loss function using the fixed weight interpolation of cross-entropy loss and KL divergence renders worse models than KL divergence alone. To leverage the training hard labels for knowledge distillation, we proposed a trust
regularization method to dynamically adjust the weight for cross-entropy loss. The experiments on \ptb dataset
showed that the student model trained via \tr got the state-of-the-art perplexity using only one third of parameters.
On the \wsj speech recognition N-best rescoring task, our knowledge distillation method reduced
the model size to $18.5\%$ of the best single model, with similar word error rates.

\bibliographystyle{IEEEbib}
\bibliography{emnlp2018}

\end{document}